\documentclass[letterpaper, 10pt, conference]{ieeeconf}      
\IEEEoverridecommandlockouts                              

\usepackage{graphics} 
\usepackage{epsfig} 
\usepackage{mathptmx} 
\usepackage{times} 
\usepackage{amsmath} 
\usepackage{amssymb}  
\usepackage{url}
\usepackage{booktabs} 
\usepackage{multirow}
\usepackage{float}
\usepackage{placeins}
\usepackage{graphicx} 
\usepackage[font=small]{caption}
\usepackage{subcaption}
\usepackage{lmodern}
\usepackage{siunitx}
\usepackage{etoolbox}
\robustify\bfseries

\usepackage{multicol}

\usepackage[ruled, shortend, linesnumbered]{algorithm2e}

\usepackage{breqn}  
\title{\LARGE \bf Estimating Motion Uncertainty with Bayesian ICP}
\author{Fahira Afzal Maken$^{1,*}$, Fabio Ramos$^{1,2}$ and Lionel Ott$^1$
	\thanks{$^*$ Corresponding author: fafz3958@uni.sydney.edu.au}
	\thanks{$^1$ School of Computer Science, The University of Sydney, Australia}
	\thanks{$^2$ NVIDIA, USA}%
}
\begin{document}
\maketitle
\thispagestyle{empty}
\pagestyle{empty}

\begin{abstract}

Accurate  uncertainty estimation associated with the pose transformation between two 3D point clouds is critical for autonomous navigation, grasping, and data fusion. Iterative closest point (ICP) is widely used to estimate the transformation between point cloud pairs by iteratively performing data association and motion estimation. Despite its success and popularity, ICP is effectively a deterministic algorithm, and attempts to reformulate it in a probabilistic manner generally do not capture all sources of uncertainty, such as data association errors and sensor noise. This leads to overconfident transformation estimates, potentially compromising the robustness of systems relying on them. In this paper we propose a novel method to estimate pose uncertainty in ICP with a Markov Chain Monte Carlo (MCMC) algorithm. Our method combines recent developments in optimization for scalable Bayesian sampling such as stochastic gradient Langevin dynamics (SGLD) to infer a full posterior distribution of the pose transformation between two point clouds. We evaluate our method, called Bayesian ICP, in experiments using 3D Kinect data demonstrating that our method is capable of both quickly and accuractely estimating pose uncertainty, taking into account data association uncertainty as reflected by the shape of the objects.
\end{abstract}

\section{INTRODUCTION}

Iterative closest point (ICP) \cite{besl_1992} is a widely used algorithm in robotics and computer vision to register two point clouds by iteratively minimizing the alignment error between them. 
ICP takes two input point clouds, source and reference, and an initial transformation estimate. 
The alignment error is minimized by repeatedly finding sets of corresponding point pairs in the two point clouds and using these to estimate a transformation that minimizes the distance between matching points. A large number of ICP variants have been proposed in the literature, building on the initial formulation proposed in \cite{besl_1992}. Many methods differ in the cost-function used, e.g. point-to-point \cite{besl_1992}, point-to-plane \cite{Chen1992}, and plane-to-plane \cite{segal2009} to name a few. These and many other variants were extensively evaluated in \cite{Pomerleau2015, Rusinkiewicz2001}.

    
    

There are several potential sources of error and uncertainty in the pose estimate obtained by ICP. For example, sensor noise and the presence of local minima in the cost function.
Another source of uncertainty comes from ambiguities in the point cloud structure, such as long featureless corridors or symmetries in objects like bowls or water bottles. This results in erroneous data associations which can cause significant estimation errors and make them impractical for localization or mapping tasks.

Current methods that estimate ICP pose uncertainty are either offline due to their high-computational cost or produce overconfident uncertainty estimates which can lead to navigation failure \cite{Iversen2017a,Bengtsson2003,Censi2007}. The reason for overconfident estimates stems from the fact that point re-associations are not taken into account in the pose uncertainty estimation step, making them unsuitable for point-to-point variants of ICP \cite{Bonnabel2016}.

A principled way to incorporate uncertainty into the ICP pose estimation process is to use a Bayesian formulation in which a prior distribution and a likelihood function are used to obtain a posterior distribution given observations. Unfortunately analytical solutions to such a Bayesian formulation are typically intractable and approximate methods have to be employed. One such approximation technique is Markov Chain Monte Carlo (MCMC) which can produce a sample-based approximation to the posterior \cite{Robert:2005}. However, traditional MCMC algorithms are still computationally demanding as they need to consider the entire dataset to produce a single sample.

In this paper we use a scalable class of MCMC techniques, called stochastic gradient Langevin dynamics (SGLD), which combines stochastic optimization with Langevin dynamics \cite{Li2016,Neal1996,Neal2010,Welling}. SGLD enables any stochastic gradient descent (SGD) based optimization to produce samples from the true posterior by injecting a controlled amount of noise into the optimization. We combine this idea with SGD-ICP \cite{fahira2018}, an ICP method that uses SGD optimization, to obtain a method which produces a pose distribution by generating samples from the posterior distribution.

The main contribution of this paper is the introduction of Bayesian ICP, an efficient and scalable ICP variant that provides a high-quality approximation over the posterior transformation. This distribution encodes both the expected pose transformation, as provided by other ICP methods, as well as uncertainty information. A C++ implementation of Bayesian ICP is available at \url{https://bitbucket.org/fafz/bayesian-icp}.

\section{RELATED WORK}

Various approaches have been proposed in the literature to estimate the pose uncertainty of ICP using closed form solutions. In \cite{Glover2013}, it is assumed that there exists a perfect one-to-one correspondence between the source and reference cloud. Under this assumption isotropic Gaussian noise is used to model the uncertainty in the position of points. It is shown that rotation errors are  distributed according to a 4-dimensional Bingham distribution.
Another way to measure uncertainty in the pose estimate relies on the Hessian of the cost function. These closed form solutions are based on the underlying assumption that the cost function of ICP can be linearized around the point of convergence. This assumption makes it possible to use a least square estimate \cite{kay, barrie} to derive the Hessian's covariance matrix analytically \cite{Bengtsson2003, Bosse2008},
or otherwise it can be approximated numerically by sampling \cite{ Biber2003}. This type of uncertainty estimation only accounts for errors which are caused by the structure of the environment but not those due to sensor noise.

A method tackling sensor noise is proposed in \cite{Censi2007}, which computes a sensor's noise error covariance as a function of an error metric in the 2D case. An extension of this idea for the 3D case is proposed in \cite{Prakhya}. This approach makes certain assumptions which may restrict their use, e.g. it is assumed that the algorithm converges to the attraction region of the true solution and that data re-association in that region does not significantly impact the objective function.


Another approach to estimate the ICP pose uncertainty is the use of sampling based methods, such as Monte Carlo algorithms \cite{Bengtsson2003,Iversen2017a}. These methods sample a large number of ICP pose transforms, requiring a complete ICP solution each time, and use the covariance of these samples as a measure of uncertainty. If a model of the environment is available, such as a computer aided design (CAD) drawing, then sensor noise can be taken into account by generating multiple scans of the environment using a sensor model \cite{Iversen2017a}. While such methods are capable of providing accurate estimates of the ICP pose uncertainty, their high computational cost makes them often unsuitable for online robotics applications. Additionally, the need for an environment model to account for sensor noise is a further limiting factor.

In this paper we propose a novel ICP method which estimates the pose distributions using stochastic gradient Langevin dynamics (SGLD). In comparison to standard MCMC based methods our method is highly efficient, joining SGD-ICP \cite{fahira2018} and SGLD to overcome the computational costs of typical MCMC approaches. In contrast to analytical methods information about the Hessian is not needed nor are assumptions made about the consistency of data associations. The general sampling based nature of the proposed approach makes the approach amenable to be used with many existing ICP variants.

\section{BACKGROUND}

In the following, we first describe the standard ICP algorithm, before giving an overview of stochastic gradient descent ICP (SGD-ICP) \cite{fahira2018}, which forms the basis of our proposed Bayesian ICP method.

\subsection{Standard ICP}

Given two input point clouds, $S = \{s_i\}_{i =1}^{N}$ and $R_{ef} =\{r_i\}_{i =1}^{M}$, where  $s_i, r_i \in \mathbb R^3$, and an initial estimate of the transformation parameters $\theta = \{x, y, z, \text{roll}, \text{pitch}, \text{yaw}\}$, the standard ICP formulation performs a rigid registration between the clouds in two steps: First, correspondences between point pairs in the two clouds are found on the basis of the Euclidean distance using the current transformation $\theta$. Second, a rigid transformation between the corresponding point pairs  is found that minimizes a cost function. These two steps are repeated until a convergence criteria, such as error threshold or iteration number, is met.

The loss $\mathcal L$ optimised by ICP in the second step is a function of $\theta$ and captures how well the transformation aligns the point pairs, i.e.:
\begin{equation}
    \operatorname*{argmin}_\theta \mathcal L(\theta) = \frac{1}{N} \sum_i^N  ||(R\; s_i+ u) - r_i||^2,
    \label{eq:icp-objective}
\end{equation}
where $u \in \mathbb R^{3 \times 1}$ is a translation vector and $R \in \mathbb R^{3 \times 3}$ is a rotation matrix.

\subsection{Stochastic Gradient Descent ICP}

SGD-ICP \cite{fahira2018} employs stochastic gradient descent (SGD) to solve the ICP optimization problem \eqref{eq:icp-objective}. In each iteration $t$ a mini-batch $\mathcal M_t$ of size $m$ is sampled from the source cloud $S$ to which the corresponding points in $R_{ef}$ are selected using standard ICP data association methods. The average gradients $\bar g$ of the parameter vector $\theta$ with respect to the loss $\mathcal L$ are computed as follows:
\begin{align}
    \bar{g} (\theta_t^{1:3}, \mathcal{M}_t) & = \frac{1}{m} \sum_i^m \Big((R_{t} \; s_i+u_t) - r_i \Big) \frac{\partial u_{t}}{\partial\theta^{1:3}_t},
    \label{eq:trans_gradient}
    \\
    \bar{g} (\theta_t^{4:6}, \mathcal{M}_t) & =  \frac{1}{m} \sum_i^m\Big(( R_{t} \; s_i+u_{t}) - r_i\Big) \frac{\partial R_{t}}{\partial \theta^{4:6}_{t}} \; s_i,
    \label{eq:rotation_gradient}
\end{align}
where $\bar g(\theta_t^{1:3}, \mathcal M_t)$ are the gradients of the translation components and $\bar g(\theta_t^{4:6}, \mathcal M_t)$ are the rotational gradients. Using this gradient information the estimate of the transformation parameters $\theta$ can be updated iteratively with the following update rule:
\begin{equation}
    \theta_{t+1} = \theta_{t} - \alpha A \bar{g} (\theta_{t}, \mathcal{M}_{t}),
    \label{eq:sgd-fomulation_simplified}
\end{equation}
where the matrix $A \in \mathbb R^{6 \times 6}$ acts as a pre-conditioner and the learning rate $\alpha$ dictates how quickly parameter values change.

By using small mini-batches, instead of the full point cloud, in each iteration SGD-ICP is computationally much more efficient than standard ICP methods. Importantly, however, the quality of the estimated transformation is equal to that of the full batch ICP approach \cite{fahira2018}.

\section{BAYESIAN ICP}

In the following we derive our proposed method Bayesian ICP, combining ideas from SGD-ICP \cite{fahira2018} and SGLD \cite{Welling} to obtain an efficient ICP method capable of estimating a pose distribution. As we will see the mini-batch based formulation of the two algorithms allows for an elegant combination of the two.

SGLD \cite{Welling} shows that adding the right amount of noise to a standard stochastic gradient descent based optimisation results in iterations converging to samples from the true posterior distribution, thus enabling Bayesian inference. As SGD-ICP is based on SGD, it is a good candidate for applying the ideas of SGLD to obtain a distribution over the transformation parameters $\theta$. To this end the general SGD-ICP update rule \eqref{eq:sgd-fomulation_simplified} is modified by adding a prior $p(\theta)$ over the transformation parameters $\theta$ and injecting Gaussian noise $\eta_t \sim \mathcal N(0,A\alpha)$ at each iteration. With these additions we obtain the following general form for Bayesian ICP:
\begin{equation}
    \theta_{t+1} = \theta_{t} - \frac{\alpha}{2} A
    \Big(
        -\nabla \log p(\theta_{t}) + N \bar{g}(\theta_{t}, \mathcal{M}_{t})
    \Big)
    + \eta_t,
    \label{eq:general_sgld_cost}
\end{equation}
where $N$ is the size of the dataset, and $\nabla \log p(\theta_{t})$ is the gradient of the log of the prior. SGLD maintains the balance between the amount of injected noise and the gradient step sizes in order to match the variance of the samples to the variance of the posterior. The convergence rates of this process have been explored in detail in \cite{Teh}.

The goal in ICP is to estimate a 6 DoF transformation which consists of three translations $\theta^{1:3}$ and three rotations $\theta^{4:6}$. The priors on these are modelled using Gaussian distributions for the translations and von Mises \cite{VON_MISES_1918} distributions for the rotations. Disregarding the normalizing constant term, the gradients of the log of  Gaussian and von Mises prior distribution are:
\begin{align}
    -\nabla_\theta \log p(\theta^{1:3}) & = (\theta^{1:3} - \mu^{1:3}) / \sigma^{1:3}, \\
    -\nabla_\theta \log p(\theta^{4:6}) & = \kappa^{4:6} \sin (\theta^{4:6}_{t} - \mu^{4:6}),
\end{align}
where $\mu^{1:6}$ represents the mean, and $\sigma^{1:3}$, and $\sigma^{4:6} = 1 / \kappa^{4:6}$ represent the variance of the prior distributions of the translation and rotation components.

Combining these priors with the SGD-ICP update rule \eqref{eq:sgd-fomulation_simplified}, and using translational and rotational gradients \eqref{eq:trans_gradient} and \eqref{eq:rotation_gradient} in the SGLD update equation \eqref{eq:general_sgld_cost} we arrive at:
\begin{multline}
    \theta^{1:3}_{t+1} = \theta^{1:3}_{t} - \frac{\alpha}{2} A
    \bigg[
        ( \theta^{1:3}_{t} - \mu^{1:3} ) / \sigma^{1:3} + \\
        \frac{N}{m} \sum_i^m \Big(
            ( R_{t} \;s_i + u_{t} ) - r_i
        \Big)
        \frac{\partial u_{t}}{\partial\theta^{1:3}_{t}}
    \bigg]
    + \eta_t^{1:3}
    \label{eq:sgd-MCMC-translation}
\end{multline}
for the translation parameters and
\begin{multline}
    \theta^{4:6}_{t+1} = \theta^{4:6}_{t} - \frac{\alpha}{2} A
    \bigg[
        \kappa^{4:6} \sin ( \theta^{4:6}_{t} -\mu^{4:6}) + \\
        \frac{N}{m} \sum_i^m \Big(
            ( R_{t} \;s_i + u_{t}) - r_i
        \Big)
        \frac{\partial R_{t}}{\partial \theta^{4:6}_{t}} \; s_i
    \bigg]
    + \eta_t^{4:6}
    \label{eq:sgd-MCMC-rotation}
\end{multline}
for the rotation parameters. As Bayesian ICP produces a collection of samples, as opposed to a single parameter vector, we can recover the posterior expectation $\mathbb{E}[\theta]$ of $\theta$ as an average over all $T$ samples, i.e. $\mathbb{E}[\theta] = \frac{1}{T} \sum_{t=1}^{T}\theta_t$.

When using a fixed step size, SGLD can exhibit slow mixing of the distribution, especially when parameters have locally different curvatures. One approach to overcome this is preconditioned SGLD \cite{Li2016} which constructs a local transform that equalizes the gradients, similar to RMSProp \cite{prop}. In this way a simple fixed step size is adequate to obtain good convergence. This gradient preconditioning is incorporated via the matrix $A$ and is computed as follows:
\begin{align}
    V(\theta_{t}) & =
        \beta V(\theta_{t-1}) +
        (1-\beta) \bar{g} (\theta_t, \mathcal{M}_t) \odot
        \bar{g} (\theta_t, \mathcal{M}_t),
    \label{eq:preconditioner}
    \\
    A(\theta_t) & =
        diag \left(
            1 \oslash \left( \lambda 1 + \sqrt{V(\theta_t) } \right) 
        \right),
    \label{eq:preconditioner_matrix}
\end{align}
where operators $\odot$ and $\oslash$ represent element-wise matrix product and division respectively. The hyper parameter $\lambda$ controls the extremes of the curvature in the pre-conditioner while $\beta \in [0, 1]$ balances the weights of the historical and current gradients.

Algorithm \ref{alg:mcmc-icp} outlines the steps performed by Bayesian ICP. First, a new mini-batch is obtained from the source cloud in line 2. This mini-batch is then transformed with the latest transformation matrix in line 3,  before in lines 4 to 8 the corresponding point pairs between the mini-batch and the reference cloud are found and stored. Once corresponding pairs have been obtained, the stochastic gradient, preconditioning, and noise are computed in lines 9 to 12. With these quantities the translation and rotation parameters are updated in line 13 and 14 respectively. Finally, after T iterations the final set of parameter samples are returned in line 16.

\begin{algorithm}[bt]
	\caption{Bayesian ICP}
	\label{alg:mcmc-icp}
	
	\SetAlgoLined
	
	\SetKwInOut{Input}{Input}
	\SetKwInOut{Output}{Output}
	\DontPrintSemicolon
	
	\Input{
		Source $S = \{s_i\}$ and reference $R_{ef} = \{ r_i \}$ clouds,
		Initial transformation parameters: $\theta_1$,
		Source cloud size: $N$,
		Mini-batch size: $m$,
		Step size: $\alpha$,
		$\lambda = 1 \times 10^{-8}$,
		$\beta = 0.9$, $V(\theta_{0}) \leftarrow 0$
	}
	\Output{Transformation samples: $\{\theta_t\}_{t=1:T}$}
	
	\For{ $t \leftarrow 1: T$}
	{
		$\mathcal{M}_{t}$ $\leftarrow $ pick a mini-batch cloud of size $m$ from $S$\;
		$\mathcal{M}_{t}$\textprime $\leftarrow$ transform mini-batch with $\theta_{t}$\;
		Pairs $ \leftarrow \emptyset$\;
		\For{$s_i^\prime \in$ $\mathcal{M}_{t}$\textprime}
		{
			$r_i \leftarrow$ closest point in $R_{ef}$ to $s_i^\prime$\;
			    Pairs $\leftarrow$ Pairs $\cup \quad \{s_i^\prime, r_i\}$\;
		}
	    \tcp{Estimate mean gradients}
	    $\bar{g} (\theta_{t} ; \mathcal{M}_t) \leftarrow  \frac{1}{m} \sum_{s_i^\prime, r_i \in \text{Pairs}} ( s_i^\prime - r_i) \frac{\partial(s_i^\prime - r_i)}{\partial \theta_{t}}$\;
		
		\tcp{Compute preconditioner and noise}
		$V(\theta_t) \leftarrow \beta V(\theta_{t-1}) + (1-\beta) \bar{g} (\theta_t; \mathcal{M}_t) \bar{g} (\theta_t ; \mathcal{M}_t)$\\
		$ A(\theta_t)\leftarrow  \bigg(1 \div \Big(\lambda + \sqrt{V(\theta_t)}\Big) \bigg)$\;
		
		$\eta_t \leftarrow \mathcal{N}(0,\alpha A(\theta_t))$\;
		
		\tcp{Update parameters}
		$\theta^{1:3}_{t+1} \leftarrow \theta^{1:3}_{t} - 0.5\alpha A(\theta_t^{1:3}) \Big( (\theta^{1:3}_{t}-\mu^{1:3})/\sigma^{1:3}+ N\bar{g} (\theta_t^{1:3} ; \mathcal{M}_t)\Big) + \eta_t^{1:3}$\;
		$\theta^{4:6}_{t+1} \leftarrow \theta^{4:6}_{t} - 0.5\alpha A(\theta_t^{4:6})\Big( \kappa^{4:6} \sin (\theta^{4:6}_{t} -\mu^{4:6})+ N\bar{g} (\theta_t^{4:6} ; \mathcal{M}_t)\Big) + \eta_t^{4:6}$\;
	
	}
	\Return{$\{\theta_t\}_{t=1:T}$}
\end{algorithm}



\section{Evaluation}
In this section we investigate the performance of Bayesian ICP in terms of the estimated pose distribution quality, impact of the number of samples, burn in length, run-time, as well as final pose estimation accuracy.

Experiments are performed on the RGB-D Scenes Dataset \cite{Kevin} which contains both partially occluded as well as a complete scans of household objects. The objects used from the dataset fall into two categories: under constrained and constrained. Under constrained objects have one or more symmetries, such as soda cans and bowls, while constrained objects have no symmetries, e.g. a coffee mug or a table. All of these objects are transformed to a known offset which is used as a ground truth pose. This dataset is selected as we expect obvious uncertainty, along the symmetry, for the under constrained objects and none for the constrained objects. 

In order to properly evaluate the quality of the estimated pose distribution we require a baseline method. For this we obtain \num{10000} transformations by running the most commonly used standard ICP to completion, each time with a different initial transformation parameters drawn from $\pm$\SI{1}{\meter} and $\pm$\SI{1}{\radian} for translation and rotation respectively. This resulting collection of ICP transformations acts as our baseline ground truth pose distribution. We additionally provide comparisons against a widely used online method \cite{Censi2007} (Online-ICP) using the point-to-plane based implementation provided by \cite{Prakhya}.

To quantify the quality of the distributions we use the Kullback-Leibler (KL) divergence between the baseline distribution and the comparison method. The final transformation quality is measured as the Euclidean distance between the estimate pose and ground truth pose for the translation, and sum of absolute angular differences for the rotation.

In the experiments our method uses a step size of \num{0.0001} with a  mini-batch size of $160$ points. To aid with convergence and regulate SGLD noise, a block decay strategy is used for the step size which is decreased by $20\%$ every \num{10} iterations towards the end of sampling. 
The parameter $N$ in \eqref{eq:general_sgld_cost}, which represents the size of the dataset, scales the mean gradient and thus impacts the choice of step size $\alpha$. As the size of point clouds varies greatly, $N$ is fixed to $N = 6000$ such that a single step size can be used. These values were coarsely picked by manual tuning. The prior mean and variance are set $0.0$ and $0.125$ respectively, however, when deployed on a platform these values could be obtained from the system's state estimation. All experiments were performed on a PC with an Intel Core i7-7700 CPU with 16GB of RAM. The algorithm was multi-threaded using OpenMP.

\subsection{Sample count}

\begin{figure}[bt]
    \centering
    
	\includegraphics[width=0.48\textwidth]{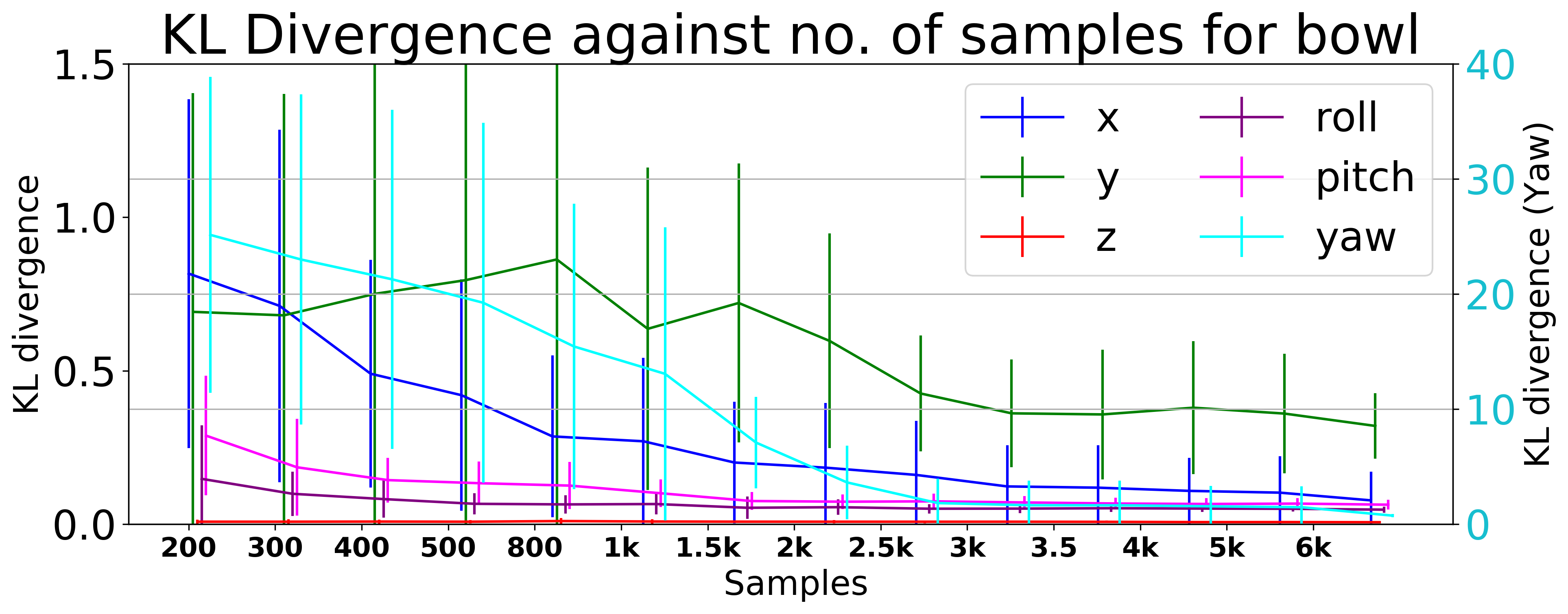}
	\includegraphics[width=0.48\textwidth]{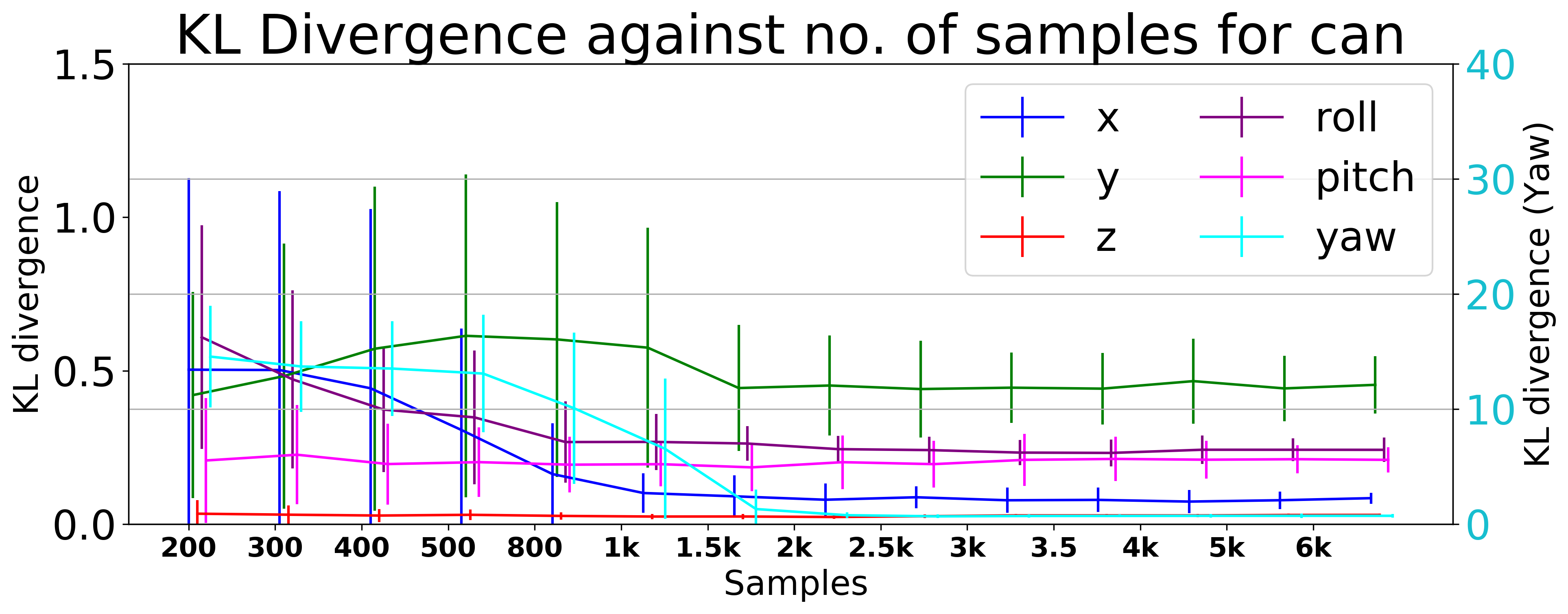}

    \caption{KL divergence of Bayesian ICP compared to the baseline for two objects bowl (top) and can (bottom), as a function of the number of samples. The majority of the improvements occur during the first \num{2000} samples.}
    \label{fig:quality_samples}
\end{figure}

We begin by evaluating the effect the number of samples has on the quality of the pose distributions estimated by Bayesian ICP. We allow Bayesian ICP to collect \num{6000} samples and repeat this ten times. Fig. \ref{fig:quality_samples} shows the evolution of mean and standard deviation of the KL divergence as the number of samples increases. The Y axis shows the KL divergence compared to the baseline, a second Y axis (right side) is used for yaw. Overall, increasing the sample size improves the quality of the distribution as both mean and standard deviation shrink. Though after a certain number of samples the iteration to iteration gains decrease. Based on these plots and other experiments \num{2000} to \num{3000} samples provide almost optimal results, and going forward we use \num{2000} samples in all other experiments.

\subsection{Burn-in length}

\begin{figure}[bt]
    \centering

	\includegraphics[width=0.48\textwidth]{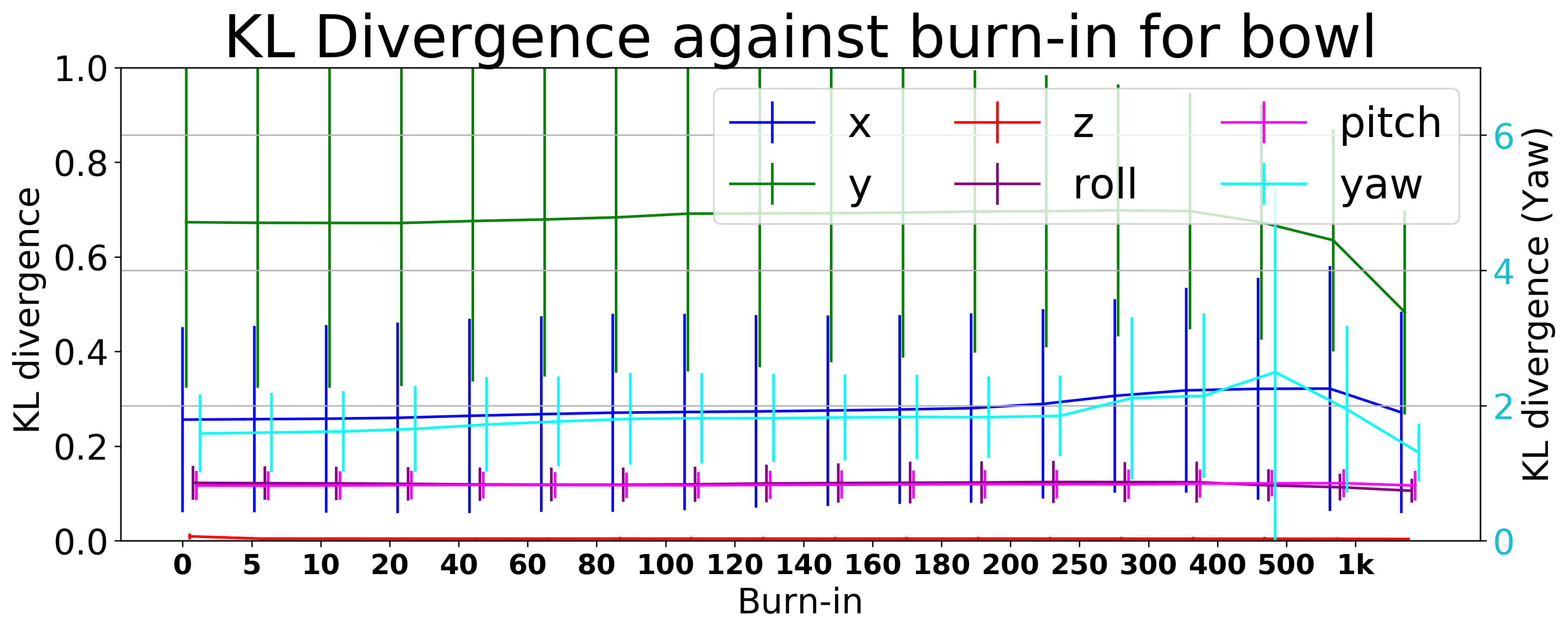}
	\includegraphics[width=0.48\textwidth]{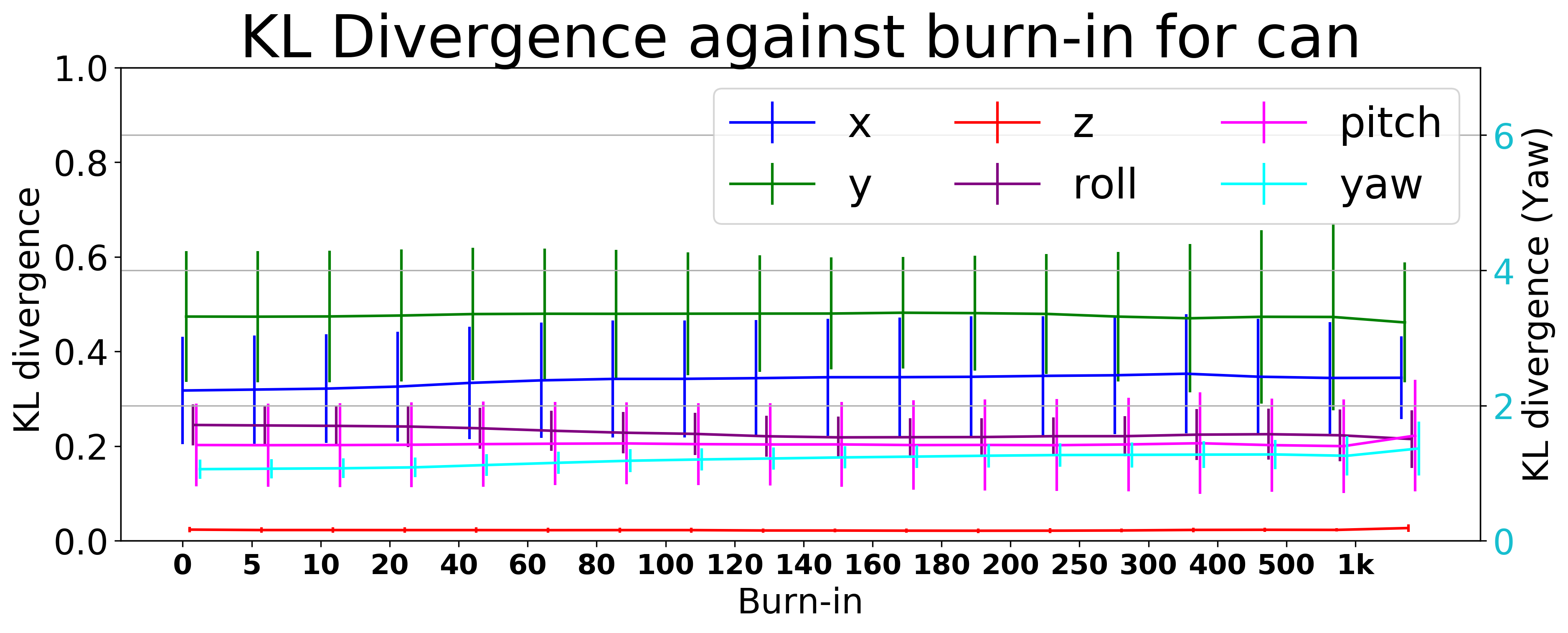}

    \caption{KL divergence of Bayesian ICP compared to the baseline for two objects bowl (top) and can (bottom), as a function of burn-in samples. Overall burn-in does not have any impact on the quality for small initial true transform.}
    \label{fig:burnin}
\end{figure}

Burn-in, which discards samples from the beginning of a sampling sequence, is common in MCMC methods and ensures that the samples produced by the scheme come from the true underlying distribution. To analyze the impact the amount of points discarded by the burn-in process has on the final distribution quality, we use the samples collected in the previous section. However, we remove a variable number of initial points and then use the subsequent \num{2000} points to obtain a distribution estimate. In Fig.~\ref{fig:burnin} we show the resulting KL divergence when compared with the baseline distribution. As we can see the size of the burn-in has no visible effect on the distribution quality. This can be explained by the fact that in the case of ICP the initial guess is close to the distribution of possible solutions and as such the sampling scheme very quickly obtains samples from the true distribution. However, for large discrepancies between initial and final transformation a certain amount of burn-in might be beneficial.

\subsection{Distribution estimation quality}

\begin{figure}[bt]
    \centering

    \begin{subfigure}{0.48\textwidth}
	    \includegraphics[page=1,width=0.49\textwidth]{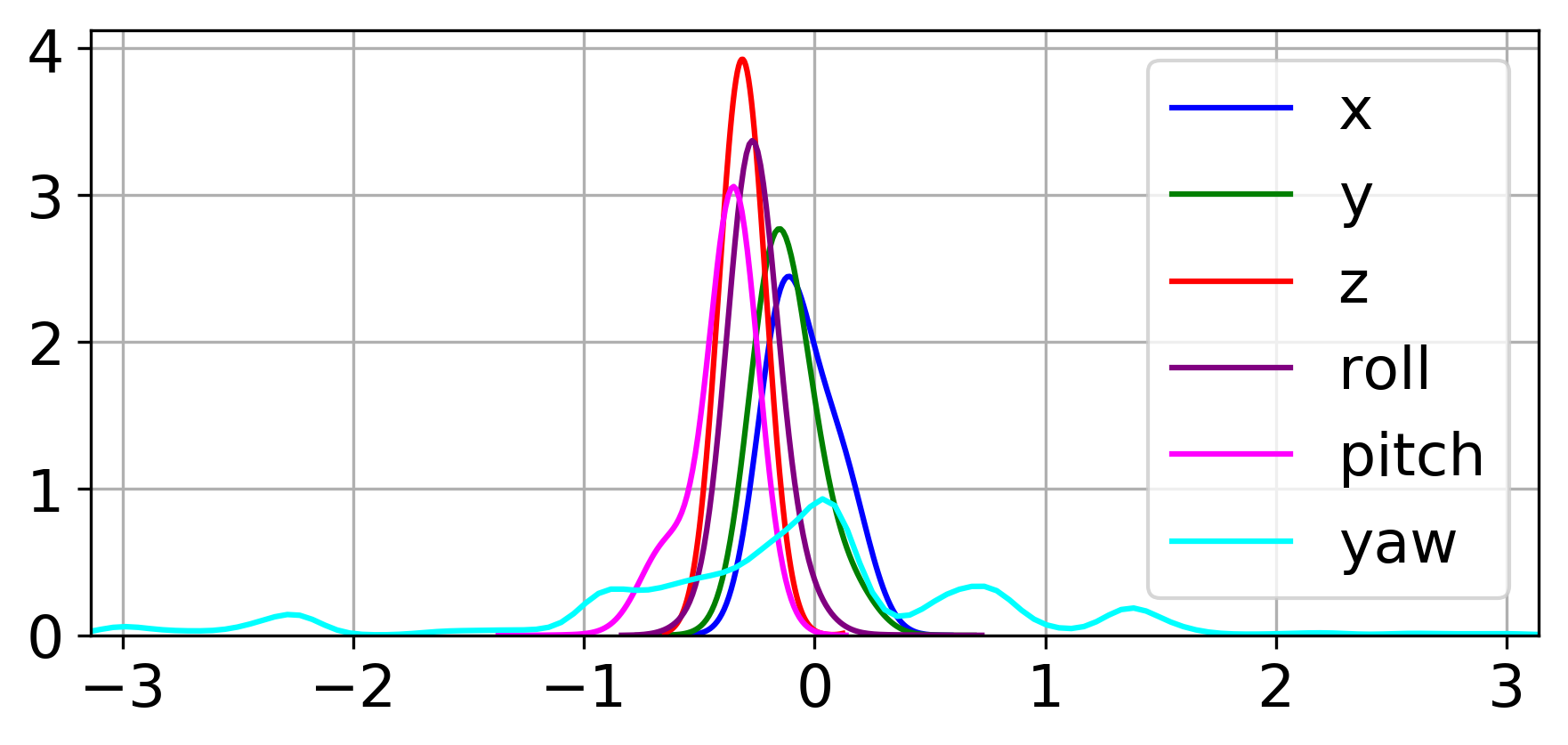}
	    \hfill
	    \includegraphics[width=0.49\textwidth]{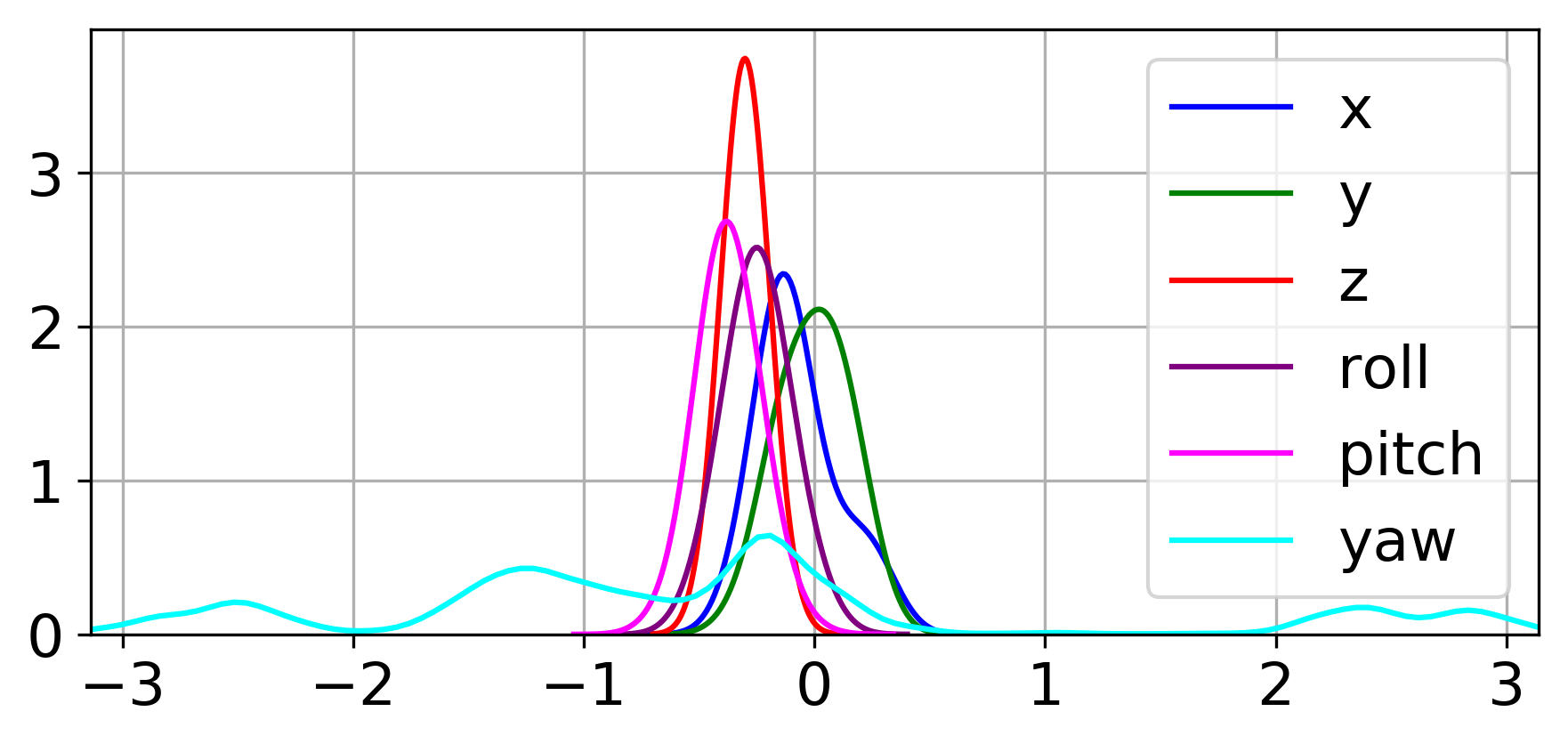}
        \caption{Bowl}
        \label{fig:kde_bowl}
    \end{subfigure}

    \begin{subfigure}{0.48\textwidth}
    	\includegraphics[page=1,width=0.49\textwidth]{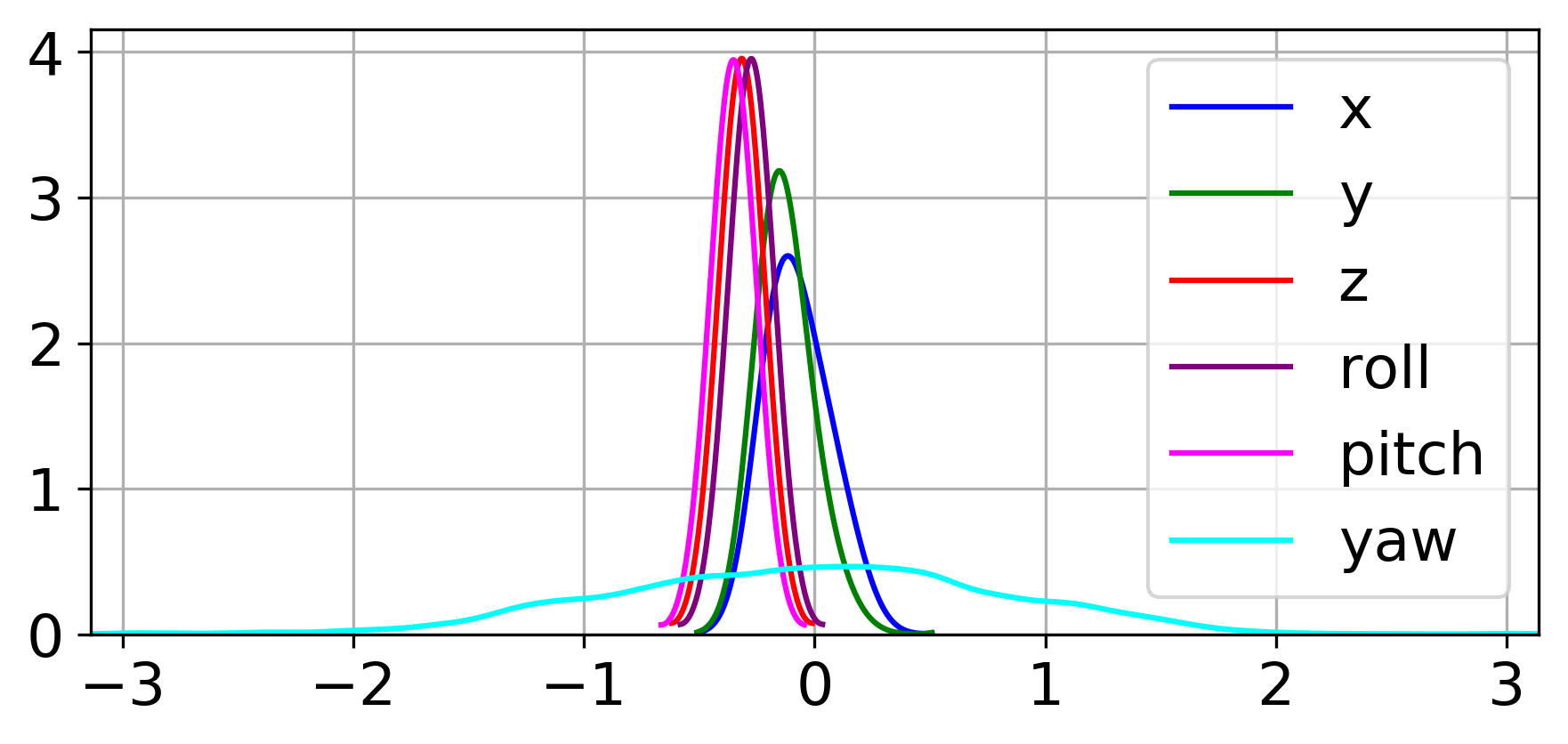}
    	\hfill
    	\includegraphics[width=0.49\textwidth]{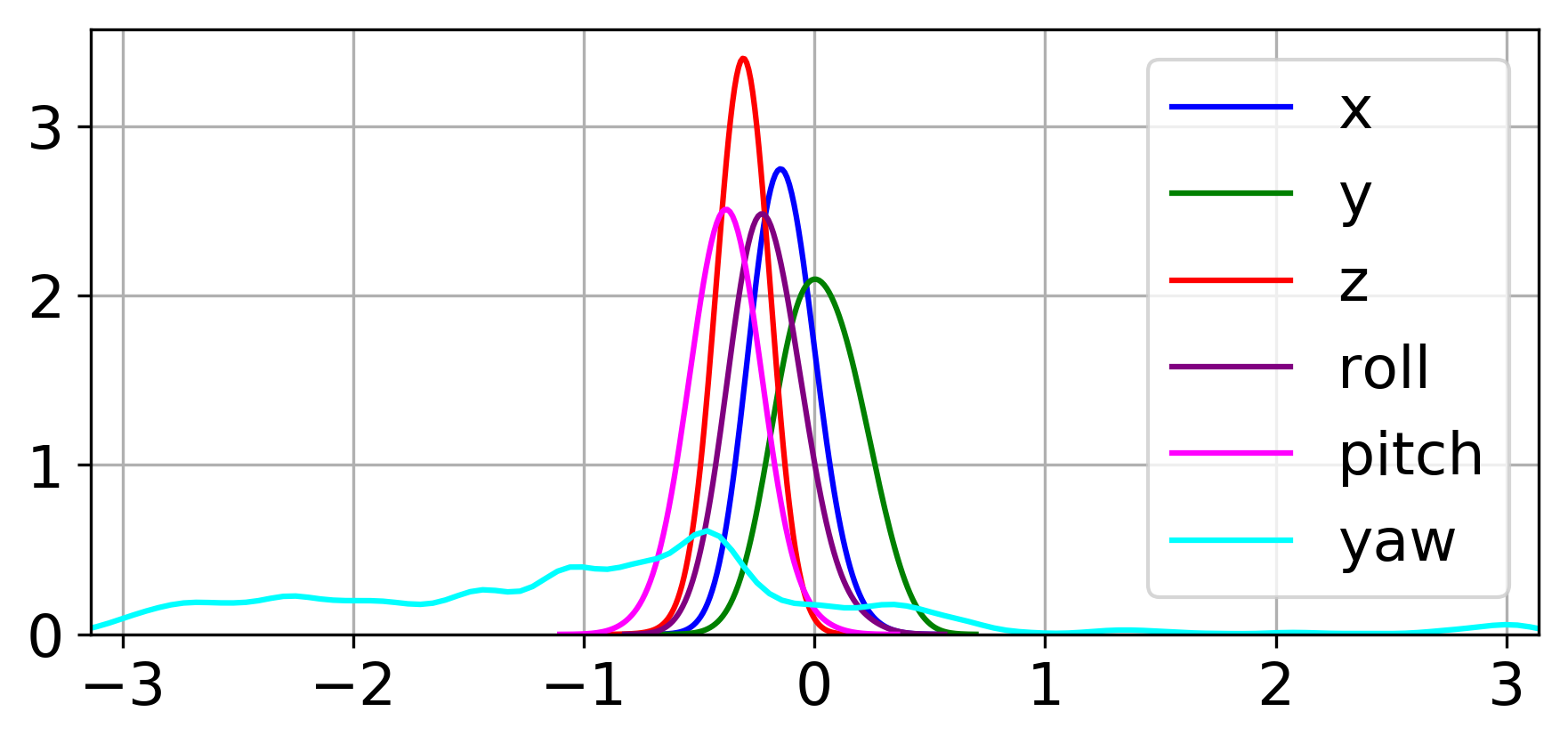}
        \caption{Soda can}
    	\label{fig:kde_can}
    \end{subfigure}
    
    
    
    
    \begin{subfigure}{0.48\textwidth}
    	\includegraphics[page=1,width=0.49\textwidth]{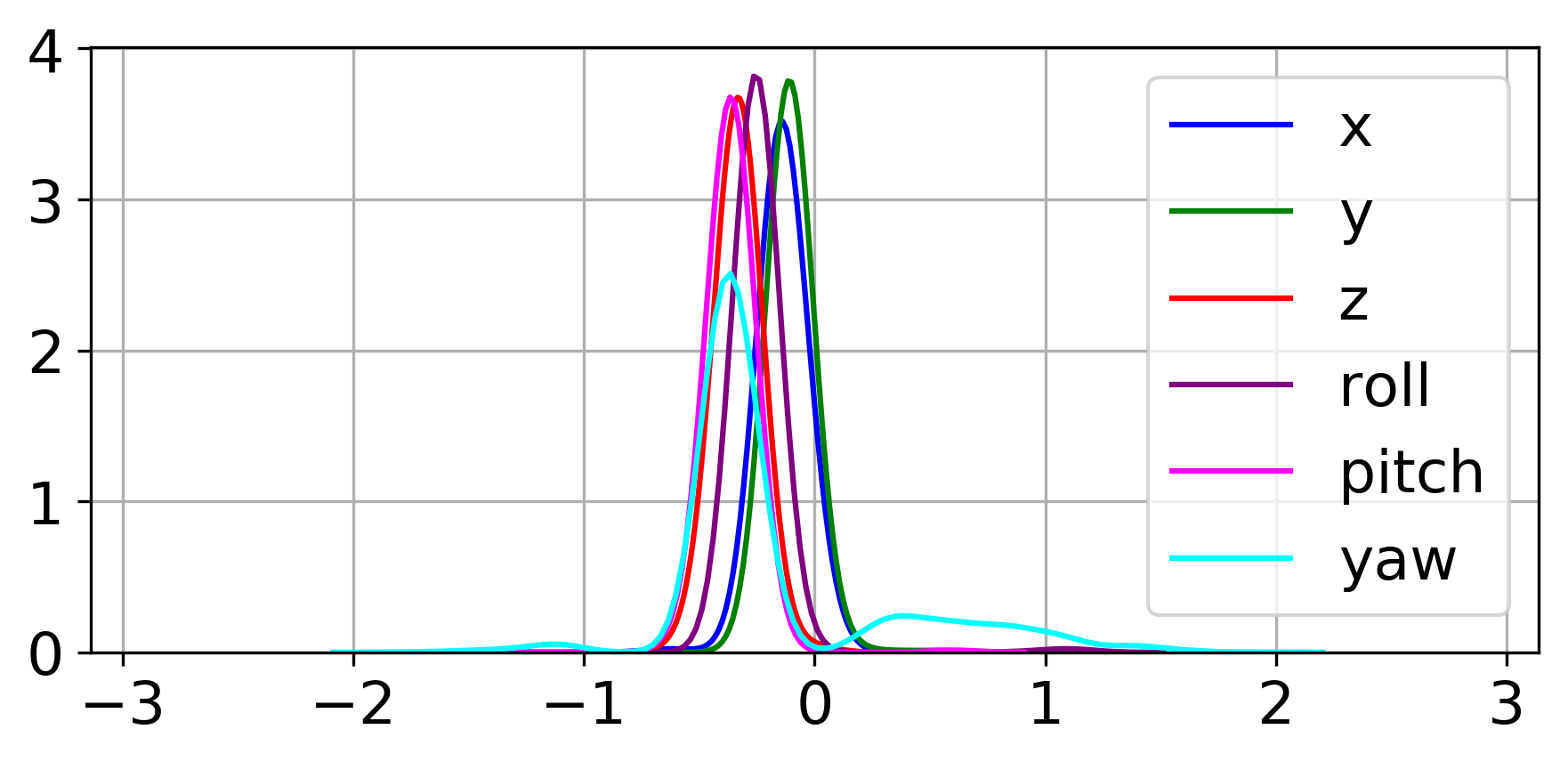}
    	\hfill
    	\includegraphics[width=0.49\textwidth]{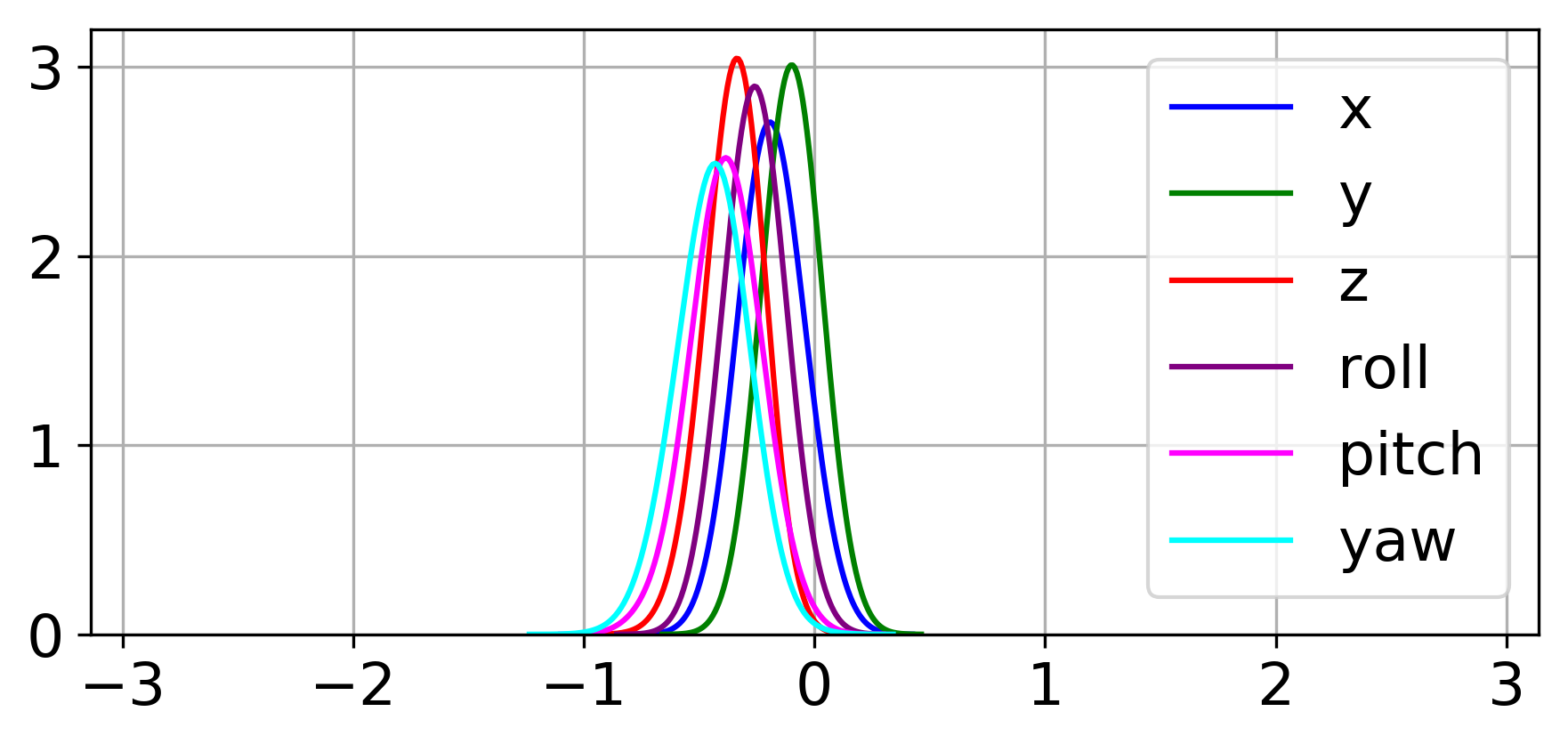}
        \caption{Mug with handle}
        \label{fig:kde_with_handle}
    \end{subfigure}


    \caption{Visualization of the pose distribution estimated by a kernel density estimator (KDE) for the baseline (left) and Bayesian ICP (right). Both bowl (a) and soda (b) can are rotationally symmetric around their yaw axis which is captured by the broad yaw distribution. This is in contrast to the mug with handle (c) which has no symmetries, resulting in all distributions being peaked.}
    \label{fig:kdes}	
\end{figure}

\begin{figure}[bt]
    \centering

    \begin{subfigure}{0.33\linewidth}
	    \includegraphics[page=1,width=\linewidth]{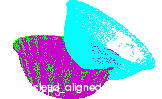}
	    \caption{Bowl}
    \end{subfigure}\hfill
    \begin{subfigure}{0.33\linewidth}
    	\includegraphics[width=\linewidth]{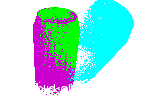}
    	\caption{Soda can}
    \end{subfigure}\hfill
    \begin{subfigure}{0.33\linewidth}
    	\includegraphics[width=\linewidth]{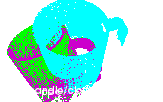}
    	\caption{Mug with handle}
    \end{subfigure}\hfill

    
    \caption{Point clouds of objects used in the experiments, (cyan) source, (magenta) reference, and (green) solution. Both (a) and (b) are under-constrained as they are symmetric around their yaw axis, while (c) has no symmetries.}
    \label{fig:clouds}	
\end{figure}

Fig. \ref{fig:kdes} shows a comparison of kernel density estimation (KDE) models of the baseline distribution (left) and the proposed Bayesian ICP method (right) on three objects. The X axis shows the transformation parameter values while the Y axis shows the density. In Fig. \ref{fig:clouds} the corresponding source scan (cyan), reference scan (magenta) and Bayesian-ICP aligned scans (green) are shown, highlighting the geometry of each object and the accuracy of the final pose. In every instances all parameters except for yaw converge close to the mean of the ground truth pose. For yaw the ground truth value is not obtained for under-constrained objects, as these are symmetric in yaw and as such all values result in the same solution. While the distribution in yaw is to some extent visually different between the baseline and Bayesian ICP solution both capture the overall picture of uncertainty. For constrained objects, which have no symmetries, the distributions show a single peak near the ground truth value for all parameters.

We provide numerical comparisons in Table \ref{tab:kl_dist} between our proposed method (Bayesian ICP) and a commonly used method \cite{Censi2007} (Online-ICP) which provides a $\mathbb R^{6 \times 6}$ covariance matrix. Bayesian ICP is run ten times, obtaining \num{2000} samples each time. The table shows the mean and standard deviation of the KL divergence between the baseline distribution and the distribution estimated by the two methods, with lower values indicating a closer match with the baseline distribution. The first four objects are symmetric around their yaw axis while the last two have no symmetries. We can see that overall our proposed method outperforms Online-ICP when it comes to matching the distribution. Especially interesting are the yaw values, where both methods tend to have higher values, however, the gap between Online-ICP and our proposed method widens significantly on objects with symmetries. This is explained by the fact that our method captures the uncertainty in the distribution as shown in previous sections, while Online-ICP tends to always predict an extremely peaked covariance matrix with values on the order of $10^{-7}$, even for yaw on objects with symmetries. This overconfident covariance estimate is also the reason that allows Online-ICP to obtain KL divergences of $0$ in cases where the baseline distribution is very peaked.

\begin{table*}[bt]
	\begin{center}
		
		\caption{Quality Comparison of the Bayesian ICP against the Online-ICP using KL Divergence}
		\label{tab:kl_dist}
	
	    \begin{tabular}{llrrrrrrr}
    		\toprule   
    		    
    	    \multirow{2}{*}{\raisebox{-\heavyrulewidth}{$\theta$}} & \multirow{2}{*}{\raisebox{-\heavyrulewidth}{Method}} & \multicolumn{7}{c}{Objects} \\

            \cmidrule{4-9}

& & & Bowl & Soda Can & Mug w/o handle & Table top & Mug w/ handle & Complete table \\
\cmidrule{1-9}

\multirow{2}{*}{x}
    & Ours && $0.387 \pm 0.199$  & $0.118 \pm 0.078$& $0.189 \pm 0.051$ & $0.619 \pm 0.044$ & $0.133 \pm 0.049$ & $0.086 \pm 0.011$ \\
    & \cite{Censi2007} && $0.731 \pm 0.000$  & $0.078 \pm 0.000$ & $0.104 \pm 0.000$ & $1.938 \pm 0.000$& $0.100 \pm 0.000$ & $0.643 \pm 0.000$

\\ \vspace{-0.75em} \\

\multirow{2}{*}{y}
    & Ours && $0.373 \pm 0.311$ & $0.163 \pm 0.168$ & $0.218 \pm 0.112$ & $1.117 \pm 0.222$  & $0.092 \pm 0.019$ & $0.082 \pm 0.016$ \\
    & \cite{Censi2007} && $0.254 \pm 0.000$ & $0.312 \pm 0.000$ & $1.596 \pm 0.000$ & $1.845 \pm 0.000$ & $0.077 \pm 0.000$ & $0.062 \pm 0.000$

\\ \vspace{-0.75em} \\

\multirow{2}{*}{z}
    & Ours && $0.007 \pm 0.003$  & $0.020 \pm 0.006$ &  $0.015 \pm 0.004$ & $0.015 \pm 0.000$ & $0.023 \pm 0.011$ & $0.008 \pm 0.001$ \\
    & \cite{Censi2007} && $0.000 \pm 0.000$ & $0.000 \pm 0.000$ & $0.055 \pm 0.000$ & $0.027 \pm 0.000$ & $0.018 \pm 0.000$ & $0.000 \pm 0.000$

\\ \vspace{-0.75em} \\

\multirow{2}{*}{roll}
    & Ours && $0.040 \pm 0.017$& $0.063 \pm 0.055$  & $0.105 \pm 0.019$  & $5.947 \pm 0.828$ & $0.425 \pm 0.022$ & $0.012 \pm 0.003$ \\
    & \cite{Censi2007} && $0.078 \pm 0.000$ & $0.000 \pm 0.000$ & $0.166 \pm 0.000$ & $7.889 \pm 0.000$ & $0.571 \pm 0.000$ & $0.000 \pm 0.000$

\\ \vspace{-0.75em} \\

\multirow{2}{*}{pitch}
    & Ours && $0.084 \pm 0.038$ & $0.051 \pm 0.045$ & $0.335 \pm 0.016$ & $1.099 \pm 0.023$ &  $0.098 \pm 0.027$  & $0.015 \pm 0.004$ \\
    & \cite{Censi2007} && $0.454 \pm 0.000$ & $0.000 \pm 0.000$ & $0.572 \pm 0.000$ & $1.568 \pm 0.000$ & $0.081 \pm 0.000$ & $0.002 \pm 0.000$

\\ \vspace{-0.75em} \\

\multirow{2}{*}{yaw}
    & Ours && $8.849 \pm 6.113$ & $6.620 \pm 3.251$ & $5.467 \pm 4.926$ & $0.842 \pm 0.254$ & $6.457 \pm 1.062$ & $1.148 \pm 0.283$ \\
    & \cite{Censi2007} && $39.689 \pm 0.001$  & $32.955 \pm 0.001$  & $370.697 \pm 0.004$ & $21.917 \pm 0.000$ & $13.438 \pm 0.001$ & $38.848 \pm 0.000$ \\

		    \bottomrule 
	    \end{tabular}
	\end{center}
\end{table*}

\subsection{Run-time}

A common concern with MCMC based methods is their runtime and as such we provide an evaluation on the runtime of our proposed method. The time needed to obtain \num{1000} samples is $8.88 \pm 3.59$~\si{\second}, where the variability comes from the size of the point clouds and number of potentially matching points in each mini-batch. However, the sampling based nature and i.i.d. assumption of MCMC allow our method to be trivially parallelized. A straight forward OpenMP based implementation improves the runtime to $6.392 \pm 0.731$~\si{\second} using two cores and $3.298 \pm 0.573$~\si{\second} using four cores. Synchronization overhead prevents perfect linear scaling between one and two cores, however, the speed gains are significant. This shows that the proposed method can produce high quality pose distribution estimates in a few seconds. For comparison, obtaining \num{1000} samples for a baseline distribution using standard ICP methods takes several hours.

\begin{figure}[bt]
    \centering
	
	\includegraphics[width=0.49\textwidth]{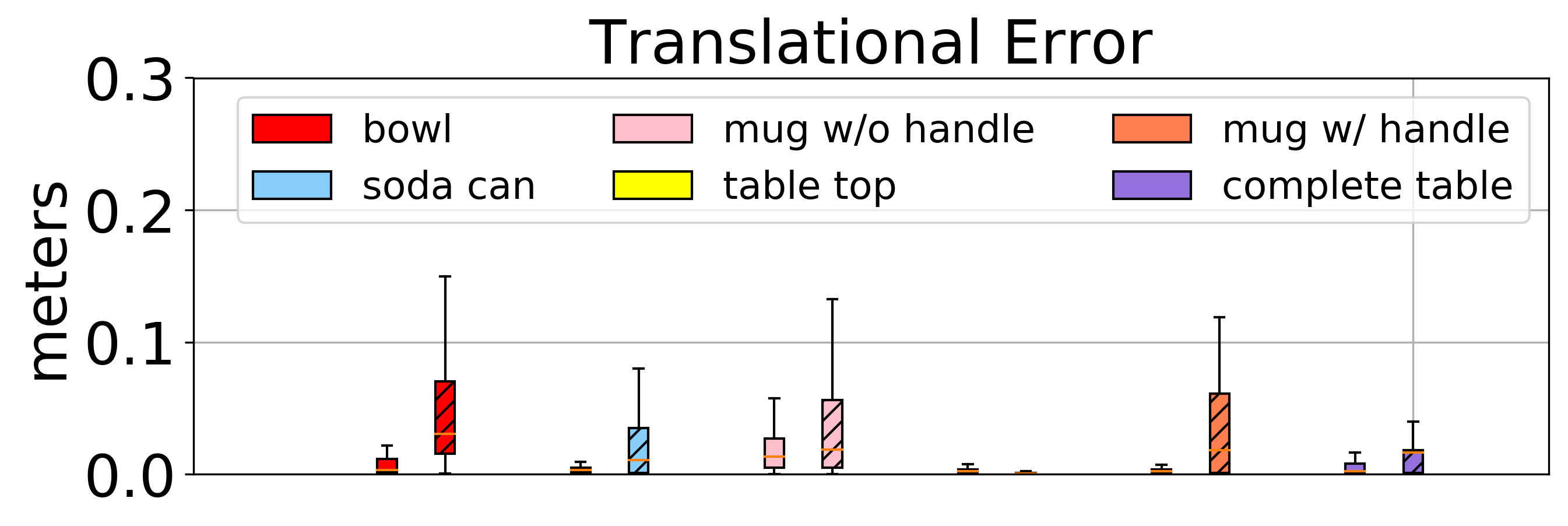}
	\includegraphics[width=0.49\textwidth]{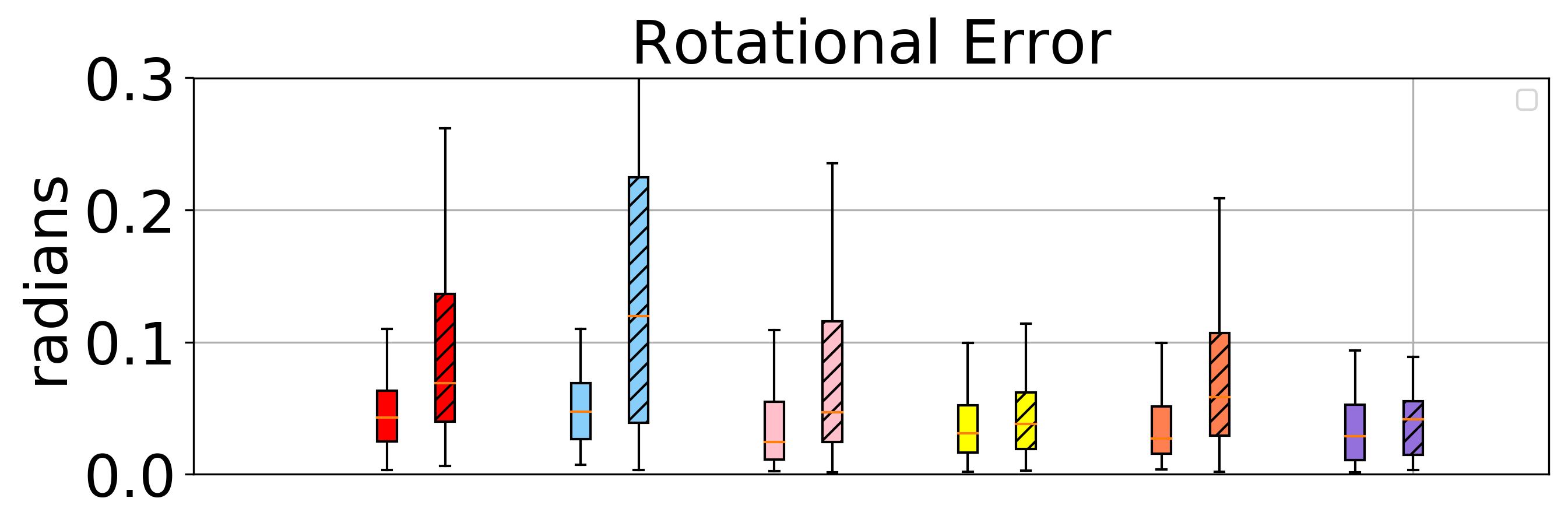}

    \caption{Side-by-side comparison of the error in translation (top) and rotation (bottom) of Bayesian ICP (left bars) and SGD-ICP (right hatched bars) for random initial true transforms over $50$ runs. Covering the solution space with samples allows Bayesian ICP to slightly improve on SGD-ICP results.}
    \label{fig:accuracy}
\end{figure}

\subsection{Pose estimation accuracy}

Finally, we investigate the quality of the final pose estimate of the proposed method and provide comparison with SGD-ICP solutions. For each object both methods start with a random initial transformation drawn from $[-1, 1]$ \si{\meter} and $[-0.3, 0.3]$ \si{\radian} for the translation and rotation component respectively. This is repeated \num{50} times and the error with respect to the ground truth transformation are shown in Fig.~\ref{fig:accuracy}, Bayesian ICP (left bars) and SGD-ICP (right hatched bars). Both methods obtain good results though in some situations Bayesian ICP obtains higher quality solutions. An explanation for this is that the sampling nature of the method forces exploration of the solution space, thus avoiding initial local minima which are selected by SGD-ICP.

\section{CONCLUSION}

This paper introduced a novel ICP method, called Bayesian ICP, which computes not only the expected transformation between two point clouds but also estimates the full pose distribution. This is achieved by leveraging the stochastic nature of SGD-ICP which is combined with stochastic gradient Langevin dynamics, an efficient MCMC method. The resulting algorithm is capable of producing high-quality posterior distributions in a few seconds compared to hours, that a standard MCMC approach would require. Extensive experiments evaluate the impact of parameter choices and showcase the ability of our method to produce accurate posterior  estimates as well as expectations of the transformation. The ability to provide high quality pose distributions has many applications in robotics from localization to registration. In future work, we plan to apply Bayesian ICP to 3D SLAM problems, integrating non-parametric filtering techniques such as particle filters into our framework. Another potential application is grasping where Bayesian ICP can be used to estimate the position of objects to be grasped, and the uncertainty estimate can be passed to a motion planner to place the gripper in a more likely position for a successful grasp.




\addtolength{\textheight}{-12cm}

\FloatBarrier

\bibliographystyle{plain}
\bibliography{bayes_icp}

\end{document}